\documentclass[letterpaper]{article} % DO NOT CHANGE THIS
\usepackage{aaai2026}  % DO NOT CHANGE THIS
\usepackage{times}  % DO NOT CHANGE THIS
\usepackage{helvet}  % DO NOT CHANGE THIS
\usepackage{courier}  % DO NOT CHANGE THIS
\usepackage[hyphens]{url}  % DO NOT CHANGE THIS
\usepackage{graphicx} % DO NOT CHANGE THIS
\usepackage{natbib}  % DO NOT CHANGE THIS AND DO NOT ADD ANY OPTIONS TO IT
\usepackage{caption} % DO NOT CHANGE THIS AND DO NOT ADD ANY OPTIONS TO IT
\usepackage{algorithm}
\usepackage{algorithmic}

\usepackage{newfloat}
\usepackage{listings}
\DeclareCaptionStyle{ruled}{labelfont=normalfont,labelsep=colon,strut=off} % DO NOT CHANGE THIS
\floatstyle{ruled}
\newfloat{listing}{tb}{lst}{}
\floatname{listing}{Listing}
\title{AnchorHOI: Zero-shot Generation of 4D Human-Object Interaction via Anchor-based Prior Distillation}
\author{
    Sisi Dai\textsuperscript{\rm 1}, Kai Xu\textsuperscript{\rm 1,2}\thanks{Corresponding author: Kai Xu, kevin.kai.xu@gmail.com}
}
\affiliations{
    \textsuperscript{\rm 1}National University of Defense Technology\\
    \textsuperscript{\rm 2}Institute of AI for Industries (IAII), Chinese Academy of Sciences (CAS)
}

\usepackage{bibentry}
\usepackage{xcolor}
\usepackage{graphicx}
\usepackage{amsmath}
\usepackage{amssymb}
\usepackage{pifont}
\usepackage{booktabs}
\usepackage{multirow}
\usepackage{makecell}
\usepackage{overpic}
\usepackage{tcolorbox}

\usepackage{color}
\definecolor{turquoise}{rgb}{0.6,0.4,0}
\definecolor{green}{rgb}{0.2, 0.8, 0.1}
\definecolor{orange}{rgb}{0.99,0.5,0.0}
\definecolor{red}{rgb}{0.9, 0, 0}
\definecolor{brown}{rgb}{0.5, 0.16, 0.16}
\definecolor{black}{rgb}{0,0,0}
\definecolor{blue}{rgb}{0,0,1}
\definecolor{teal}{rgb}{0.0, 0.4, 0.4}
\definecolor{purple}{rgb}{0.65,0,0.65}
\definecolor{dark_green}{RGB}{10,147,157}
\definecolor{yellow}{RGB}{238, 155, 0}
\definecolor{dark_red}{RGB}{174, 32, 17}

\newcommand{\ssd}[1]{{\color{black}\textbf{}#1}\normalfont}
\newcommand{\prior}[1]{{\color{black}\textbf{}#1}\normalfont}

\newcommand{\delete}[1]{}

\usepackage{xspace}
\newcommand{\nickname}{AnchorHOI}

\begin{document}

\maketitle

\begin{abstract}

Despite significant progress in text-driven 4D human-object interaction (HOI) generation with supervised methods, the scalability remains limited by the scarcity of large-scale 4D HOI datasets. To overcome this, recent approaches attempt zero-shot 4D HOI generation with pre-trained image diffusion models. However, interaction cues are minimally distilled during the generation process, restricting their applicability across diverse scenarios. In this paper, we propose AnchorHOI, a novel framework that thoroughly exploits hybrid priors by incorporating video diffusion models beyond image diffusion models, advancing 4D HOI generation. Nevertheless, directly optimizing high-dimensional 4D HOI with such priors remains challenging, particularly for human pose and compositional motion. To address this challenge, AnchorHOI introduces an anchor-based prior distillation strategy, which constructs interaction-aware anchors and then leverages them to guide generation in a tractable two-step process. Specifically, two tailored anchors are designed for 4D HOI generation: anchor Neural Radiance Fields (NeRFs) for expressive interaction composition, and anchor keypoints for realistic motion synthesis. Extensive experiments demonstrate that AnchorHOI outperforms previous methods with superior diversity and generalization. 
%Code will be released to facilitate further research in 4D HOI generation.

\end{abstract}

%% import symbols
\newcommand{\renderer}{g}
\newcommand{\modelparams}{\Phi}
\newcommand{\renderedimage}{x}
\newcommand{\camerapose}{\pi}
\newcommand{\textprompt}{T}
\newcommand{\textembedding}{y}
\newcommand{\timestep}{t}
\newcommand{\noise}{\epsilon}
\newcommand{\noisepredictnet}{\phi} %from T2I diffusion model
\newcommand{\bodypose}{\theta}
\newcommand{\bodyshape}{\beta}
\newcommand{\spatialpoint}{p}
\newcommand{\allpoints}{P}
\newcommand{\crossentropy}{CE}
\newcommand{\hyperparameter}{\eta}
\newcommand{\distancetoanchor}{d}

%% The actual document with your content starts here

\section{1\quad Introduction}

\label{sec:intro}

Humans constantly interact with surrounding objects in daily life, like sitting on a chair, carrying a backpack, or playing a guitar. Text-driven generation of 4D human-object interaction (HOI) is foundational to the 4D virtual world, and has garnered increasing attention for its potential in applications such as AR/VR, video games, embodied AI, and robotics, among many others.

However, generating realistic 4D HOI from natural language remains a challenging task, \prior{as it requires extensive prior knowledge to understand} both the inherent spatio-temporal complexity and the broad spectrum of interaction types. Existing approaches~\cite{bhatnagar22behave, diller2024cg, li2024controllable} primarily follow the supervised learning paradigm, relying on paired text-HOI data~\cite{li2023object, bhatnagar22behave, jiang2023full} as ground-truth \prior{to learn such priors}. However, collecting such data at scale is difficult and costly, as it demands sophisticated motion-capture (mocap) for both humans and objects, along with labor-intensive annotations. The limited scale of existing data heavily constrains the scalability and diversity of these supervised approaches. 

\ssd{Recent approaches have taken initial steps toward the zero-shot learning paradigm, aiming to eliminate reliance on paired text-HOI data. While efforts like InterDreamer~\cite{xu2024interdreamer} remove the need for paired text annotations, they still rely on mocap-based HOI data. More recently, methods such as AvatarGO~\cite{cao2024avatargo} attempt to substitute mocap-based data by \prior{distilling priors from a pre-trained image diffusion model}. However, they focus solely on relative HOI positioning, leaving human deformation and interactive motion with objects unaddressed: (i) the human remains fixed in a canonical pose during interaction composition; (ii) the motion source is derived from a text-to-human motion
 model without object awareness. While these overlooked aspects are indispensable for realistic 4D HOI generation, they remain unexplored due to the inherent complexity. This calls for both \prior{richer priors} beyond image diffusion models and more advanced prior distillation techniques for effective guidance.}

To this end, we propose \nickname{}, a novel framework that exploits \prior{hybrid priors from pre-trained image and video diffusion models}. Given a natural language description as input, \nickname{} (i) first compose interactions by deeply exploring priors from image diffusion models; (ii) then synthesize motion by leveraging rich motion priors learned from video diffusion models, without the reliance on mocap-based data for either human interaction or motion. 

\ssd{However, achieving expressive interaction composition and realistic motion synthesis is far from straightforward, as two key challenges still stand in the way.} \textbf{\textit{(i) Adaptive human pose optimization \prior{under image diffusion models}.}} Existing approaches typically fix the human pose during composition, lacking adaptability to interaction-specific scenarios. While adaptive human pose optimization is essential for composing expressive interactions, the complex articulated structure of the human body leads to a high degree of freedom, making such optimization under diffusion priors challenging. \textbf{\textit{(ii) Compositional motion extraction \prior{from video diffusion models}.}} While recent video diffusion models demonstrate strong capabilities in generating realistic and diverse motion sequences, they often exhibit inter-subject occlusions in compositional scenarios, precisely where grounded human-object contacts occur. This makes it challenging to extract reliable interaction-aware motion for both human and object subjects.

To address these challenges, we introduce a novel anchor-based prior distillation strategy, which circumvents the difficulty of direct optimization under diffusion models. Specifically, through a tractable two-step process: first constructing interaction-aware anchors from textual descriptions, and then leveraging them to guide the target generation. With two tailored anchors, our \nickname{} incorporates two key innovations as follows: (i) \textbf{interaction composition via anchor NeRF}. NeRF, a more effective representation for distilling interaction priors from image diffusion models than complex parametric human models, is thus adopted as our anchor bridge. To alleviate NeRF noise and enhance semantic consistency, we perform pose alignment between the skeletons of the desired human avatar and the anchor NeRF. By composing the posed human avatar with the target object, we achieve thoroughly distilled HOI generation. 2) \textbf{motion synthesis via anchor keypoint}. Purely visual cues often miss essential interaction motion information due to occlusion, where keypoints fortunately lie. Consequently,  occluded contact keypoints and body keypoints are well suited as anchors. With these anchor keypoints, occluded motion information is reliably recovered, enabling interaction‑faithful 4D HOI synthesis.

% 总结contributions
Our contributions are summarized as follows:

\begin{itemize} 
\item \nickname{} takes a further step toward zero-shot text-driven 4D HOI generation, thoroughly exploiting hybrid priors by pioneering anchor-based prior distillation.
\item By leveraging anchor NeRFs and anchor keypoints for static interaction composition and dynamic motion synthesis, \nickname{} circumvents the challenges of high-dimensional optimization and achieves expressive 4D HOI generation.
\item Extensive qualitative and quantitative evaluations show that our \nickname{} substantially outperforms existing methods in both static 3D and dynamic 4D HOI generation. 
\end{itemize}

\section{2\quad Related Work}

\subsection{3D Content Generation}
Benefiting from advances in diffusion-based text-to-image generation~\cite{saharia2022image, saharia2022imagen, gu2024filter, huang2019enhancing}, DreamFusion~\cite{poole2022dreamfusion} introduced Score Distillation Sampling (SDS) for text-to-3D generation using NeRF, by distilling guidance from pre-trained diffusion models. Subsequent works have improved output quality~\cite{lin2023magic3d, wang2023prolificdreamer}, controllability~\cite{metzer2022latent-nerf}, and efficiency~\cite{wu2024consistent3d}, while also exploring textured reconstruction~\cite{richardson2023texture, cao2023dreamavatar}.  
For 3D humans, methods such as~\cite{kolotouros2023dreamhuman} generate controllable avatars, though they often require input-specific optimization.  
Recent approaches like Zero123++~\cite{shi2023zero123++} and MVDream~\cite{shi2023MVDream} leverage 2D diffusion models to synthesize consistent multi-view images, serving as inputs for efficient 3D reconstruction~\cite{liu2023syncdreamer}.  Large reconstruction models~\cite{hong20243dtopia, xu2023dmv3d} further scale this direction by adopting transformer-based architectures.  Despite these advances, generating complex, compositional 3D scenes remains a significant challenge.

\subsection{3D Compositional Generation}
To address the challenge of compositional 3D generation, recent works have explored object layout and relational reasoning. Epstein et al.~\cite{epstein2024disentangled} and GALA3D~\cite{zhou2024gala3d} optimize component arrangements for multi-object scenes. ComboVerse~\cite{chen2024comboverse} introduces spatial-aware SDS to model relations, while GraphDreamer~\cite{gao2023graphdreamer} leverages large language models to construct object-relation graphs. Despite this progress, modeling human-object interactions remains underexplored. Recently, InterFusion~\cite{dai2024interfusion} generates human-object scenarios by retrieving human poses from offline-constructed, image-reconstructed pose datasets. However, the retrieved poses remain fixed during optimization, limiting adaptability to specific interaction contexts.

\subsection{4D Content Generation}
Recent progress in video diffusion models~\cite{gu2023factormatte} and score distillation sampling has advanced diverse approaches for 4D scene generation. Make-A-Video3D~\cite{singer2022make-a-video} adopts HexPlane features for 4D representation. 4D-fy~\cite{bahmani20234d} and DreamGaussian4D~\cite{ren2023dreamgaussian4d} use multi-stage pipelines to animate static 3D content. Dream-in-4D~\cite{zheng2023unified} supports personalized 4D generation via image guidance, while Consistent4D~\cite{jiang2023consistent4d} synthesizes scenes from video input using RIFE~\cite{huang2022real} and super-resolution. 4DGen~\cite{yin20234dgen} and AnimatableDreamer~\cite{wang2023animatabledreamer} enable controllable motion via driving videos. More recently, Comp4D~\cite{xu2024comp4d} and TC4D~\cite{bahmani2024tc4d} introduce trajectory-based generation for compositional 4D scenes. Despite these advances, generating 4D human avatars with realistic object interaction remains challenging. The recent approach AvatarGO~\cite{cao2024avatargo} attempts to address this; however, it lacks human articulation modeling during interaction composition, leading to limited interaction outcomes, such as simple holding.

\section{3\quad Preliminary Knowledge}
%For brevity, we introduce only SDS here and refer to the Supplement for NeRF~\cite{mildenhall2020nerf} and SMPL-X~\cite{SMPL-X:2019} details.

\paragraph{SDS.} Score Distillation Sampling (SDS), introduced in DreamFusion~\cite{poole2022dreamfusion}, performs iterative optimization to align 3D representations with text-to-image diffusion priors.  Compared to non-iterative image generation followed by reconstruction, SDS more reliably distills semantics encoded in image diffusion models, particularly for complex interactions. While $\renderedimage=\renderer(\modelparams)$, $\renderedimage$ is the 2D image rendered by a differentiable renderer $\renderer$ with model parameters $\modelparams$ (\emph{e.g.} the MLPs correspondingly in NeRF), under a randomly sampled camera pose. By injecting the sampled noise $\noise$ into $\renderedimage$ at a time step $\timestep$, the noisy image $\renderedimage_{\timestep}$ is produced. The pre-trained 2D text-to-image diffusion model ${\noisepredictnet}$ provides a denoising network $\hat{\noise}_{\noisepredictnet}(\renderedimage_{\timestep} ; \textembedding, \timestep)$ that predicts the noise $\hat{\noise}$ given the noisy image $\renderedimage_{\timestep}$, time step $\timestep$, and text embedding $\textembedding$. SDS then optimizes the model parameters ${\modelparams}$ by minimizing the difference between the predicted noise and the added noise:
\begin{equation}
\nabla_{\modelparams} \mathcal{L}_{\mathrm{SDS}}(\noisepredictnet, \renderedimage) = \mathbb{E}_{\timestep, \noise} [w(\timestep) (\hat{\noise}_{\noisepredictnet}(\renderedimage_{\timestep}; \textembedding, \timestep)-\noise) \frac{\partial \renderedimage}{\partial \modelparams}],
\end{equation}
where $w(\timestep)$ is the weighting term at the time step $\timestep$.

\section{4\quad Method}

In this section, we begin with problem formulation, followed by anchor illustration and pipeline overview. We then detail our two core components: (i) interaction composition via anchor NeRF, and (ii) motion synthesis via anchor keypoint. 

\subsection{Problem Formulation}
\label{subsec:illustration}

AnchorHOI aims to generate dynamic 3D HOI sequences $\mathcal{X}_d$, conditioned on textual input $T$, \emph{i.e.} $ T \rightarrow \mathcal{X}_d$. The input is a natural language description, denoted as $T = \{T_{\textrm{inter}}, (T_{\textrm{motion}})\}$, where $T_{\textrm{inter}} = \{T_{\textrm{human}}, T_{\textrm{action}}, T_{\textrm{object}}\}$ specifies the desired human avatar, interaction type, and object category, respectively. $T_{\textrm{motion}}$ is optional to provide a more detailed motion description. The output is a sequence of 3D HOIs, $\mathcal{X}_d = \{(H_i, O_i)\}_{i=0}^{L-1}$, where $H_i$ and $O_i$ denote the human and object representations at frame $i$, and $L$ is the total number of frames.

The human representation is defined as $H_i = \{s, r_i^h, t_i^h, M(\theta_i, \Theta)\}$, where $s$ denotes relative scale to the object, $r_i^h$ and $t_i^h$ denote the global rotation and translation, respectively. $M(\theta_i, \Theta)$ represents the human avatar animated by the articulation pose $\theta_i$, which is an explicit mesh defined as $M(\theta_i, \Theta) = \{V, F, C\}$. The posed vertices $V = \mathcal{M}(\theta_i, \psi, \beta, \mathbf{D})$ and faces $F$ are given by the parametric human body model SMPL-X~\cite{SMPL-X:2019}, and $C$ represents the vertex colors. We denote $\Theta = (\beta, \mathbf{D}, C)$, the parameters for shape sculpting and appearance generation. The object representation is defined as $O_i = \{r_i^o, t_i^o, \Phi\}$, where $r_i^o$ and $t_i^o$ denote the global rotation and translation, $\Phi$ represents the object identity including both geometry and appearance.

Following state-of-the-art methods~\cite{cao2024avatargo}, we first generate a static 3D HOI instance \(\mathcal{X}_s = (H_s, O_s)\), and subsequently extend it to a dynamic 3D HOI sequence \(\mathcal{X}_d\), \emph{i.e.}, \(T \rightarrow \mathcal{X}_s \rightarrow \mathcal{X}_d\).

\subsection{Anchor illustration and Pipeline Overview}
\begin{figure}[!t]
    \centering
    \includegraphics[width=1\linewidth]{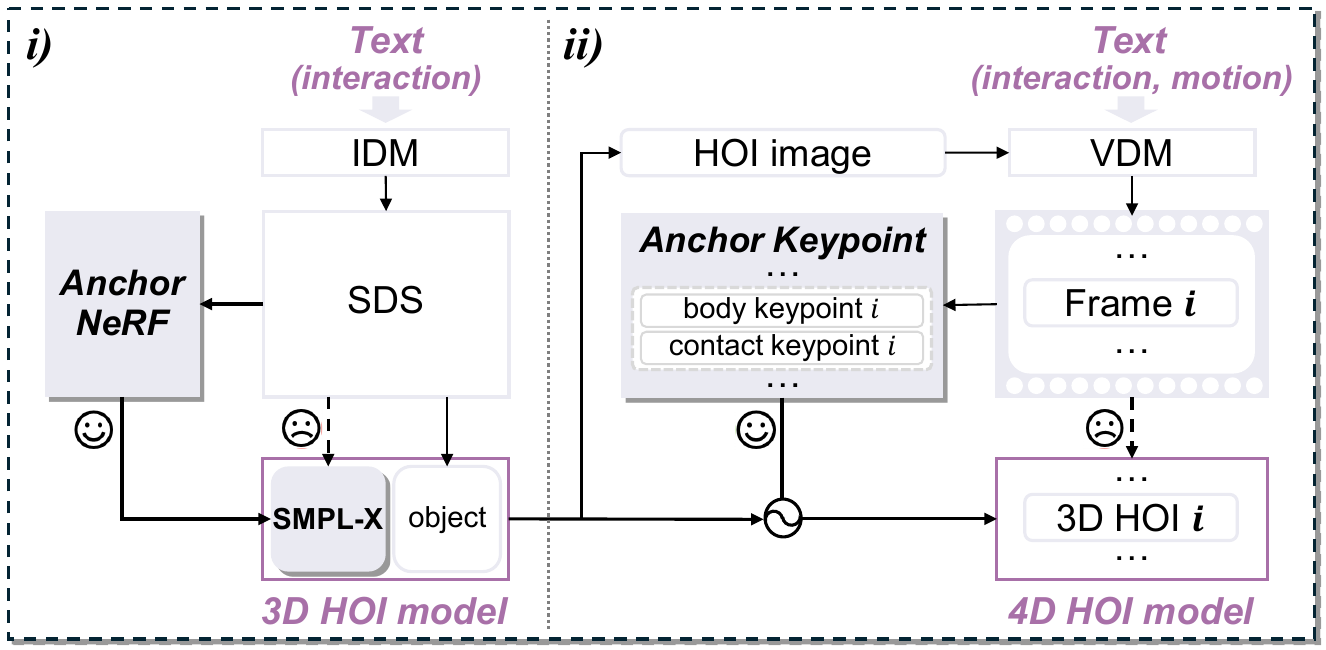}
    %\vspace{-0.5em}
    \caption{Anchor illustration.}
    \label{fig:anchor_illustration}
    %\vspace{-1.5em}
\end{figure}
\paragraph{Anchor illustration.}
As illustrated in Figure \ref{fig:anchor_illustration}, directly distilling priors from image and video diffusion models (IDMs and VDMs) for zero-shot 4D HOI generation, faces two key challenges: \textit{(i)} integrating interaction priors from IDMs into the SMPL-X model (interaction), and \textit{(ii)} transferring 2D interaction motion from VDMs for HOI. Unfortunately, previous work on zero-shot 4D HOI generation avoids these challenges due to the inherent difficulty. 

We therefore propose an anchor-based strategy, introducing intermediate anchors to bridge SMPL-X and HOI motion with the priors from IDMs and VDMs. Specifically, two anchors are tailored:

(i) Anchor NeRF. As a visual-prioritized neural representation, NeRF is more adept at distilling rich visual priors from IDMs than the geometry-prioritized SMPL-X model, and thus an ideal anchor bridge. However, visual results from NeRF often suffer from noise and struggle to be directly transferred to the SMPL-X model. We therefore turn to skeleton information rather than pixel-level visual information. Specifically, by aligning the SMPL-X pose with detected skeletons, AnchorHOI enables effective optimization of the SMPL-X structure, thereby capturing reliable interaction cues. 

\begin{figure*}
    \centering
    \includegraphics[width=0.95\textwidth]{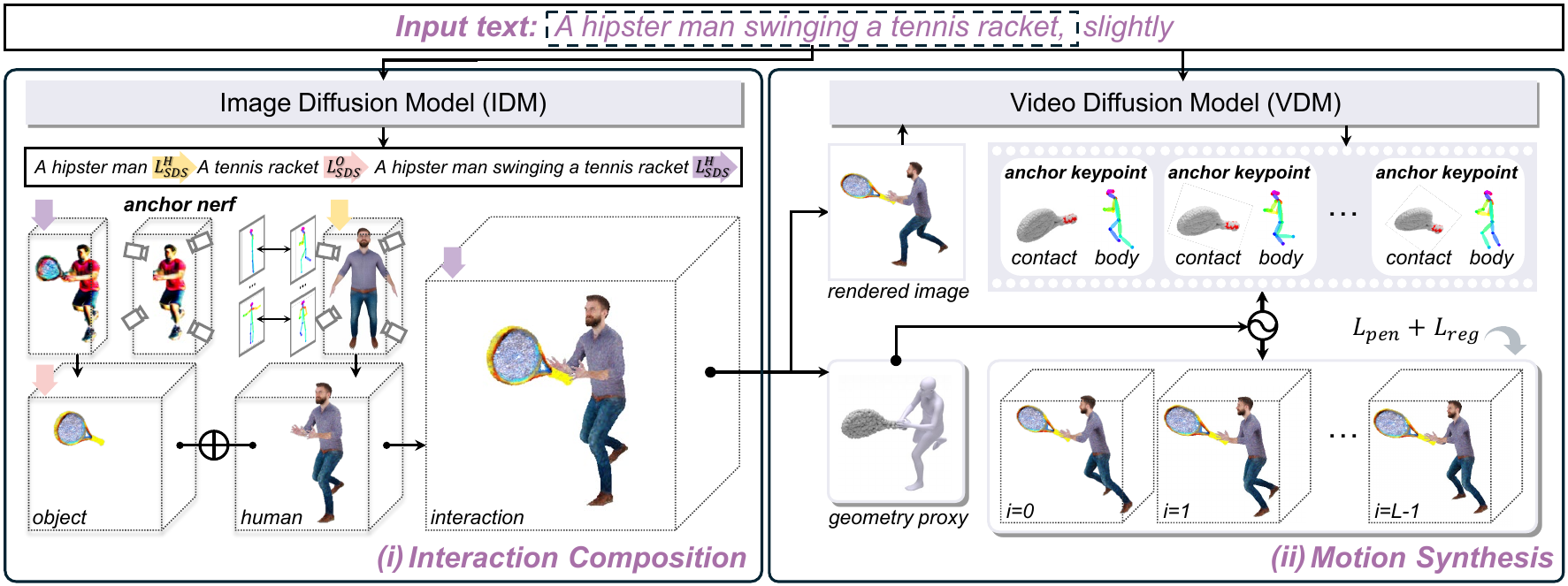}
    %\vspace{-0.7em}
\caption{Pipeline overview of AnchorHOI, consisting of two components: (i) interaction composition and (ii) motion synthesis.}
\label{fig:overview}    
%\vspace{-1em}
\end{figure*}

(ii) Anchor keypoints. Visual results of videos generated from VDMs often suffer from inter-subject occlusions in compositional HOI scenarios, missing essential inter-frame differences. In contrast, motion keypoints deliver more robust interaction cues, serving as an ideal anchor bridge. A natural question then arises: \textit{What keypoints are customized anchors for HOI?} To this end, we define anchor keypoints as a combination of body and contact keypoints, offering coherent tracking cues, thus providing a simple yet effective solution for reliable 4D HOI synthesis.

\paragraph{Pipeline overview.}
As illustrated in Figure~\ref{fig:overview}, the \nickname{} pipeline comprises two sequential components: (i) interaction composition. We adopt NeRF representations to exploit priors from IDMs and extract human part as anchor NeRF. SMPL-X poses are then optimized to align with skeletons detected from the anchor NeRF, thus capturing interaction cues that are challenging to obtain directly from IDMs. (ii) motion synthesis. We extract both body and contact keypoints from the motion generated by VDMs, treating them as anchor keypoints. The 3D HOI sequences are subsequently optimized to track these anchor keypoints, thereby capturing interaction cues that are otherwise difficult to extract directly from VDMs.

% \subsection{Static Interaction Composition} 
\subsection{Interaction Composition} 
\label{subsec:Interaction}
This part aims to generate a static 3D HOI instance $\mathcal{X}_s$, \emph{i.e.} $ T_{\textrm{inter}} \rightarrow (H_s, O_s)$ with anchor NeRF based prior distillation from pre-trained IDMs. We first introduce \textit{(i)} \emph{Anchor NeRF Generation}, and then \textit{(ii)} \emph{Interaction Generation with Anchor NeRF}.

\paragraph{Anchor NeRF generation.}
We construct an anchor NeRF from a coarse, entangled human-object NeRF optimized under the guidance of IDMs. Specifically, we first obtain a NeRF representation \(\hat{\Phi}\)  aligned with the textual prompt $T_{\textrm{inter}}$ via SDS optimization: $\nabla_{\hat{\Phi}} \mathcal{L}_{\mathrm{SDS}}(\hat{\renderedimage})$. We then extract human-isolate NeRF from \(\hat{\Phi}\), serving as the anchor NeRF. The extraction is conducted using a multi-view feature alignment loss, formulated as a mean squared error between features: \(\nabla_{\hat{\Phi}_A} \mathcal{L}_{\mathrm{align}}(\mathcal{F}(\renderedimage), \mathcal{F}(\hat{\renderedimage}))\), where \(\mathcal{F}(\cdot)\) denotes the masked RGB feature extractor here.

% from object interference

\paragraph{Interaction generation with anchor NeRF.}

For human $H_s$ generation, we generate the posed human avatar $H_s = \{s, r_{s}^h, t_{s}^h, M(\theta_s, \Theta)\}$ in a two-step manner.  
To ensure structural completeness and identity consistency with $T_{\textrm{human}}$, we first generate a canonical avatar using standing pose \(\theta_{\textrm{stand}}\), optimizing shape and appearance \(\Theta\) via SDS optimization: \(\nabla_{\Theta} \mathcal{L}_{\mathrm{SDS}}(\renderedimage)\). This generation process incorporates recent advanced techniques of human avatar generation methods. %We refer readers to the Supplementary Material for details.

We then fit the generated 3D animatable human avatar to the anchor NeRF \(\hat{\Phi}_{A}\), to optimize the remaining parameters \(\{s, r_s^h, t_s^h, \theta_s\}\). Specifically, we project canonical views of the anchor NeRF and extract 2D skeleton keypoints using OpenPose~\cite{cao2019openpose}. We then minimize the discrepancy between the projected 3D SMPL-X joints and the detected 2D keypoints across multiple views:
\begin{equation}
    \label{eq_joint_fit}
    \nabla_{\{s, r_s^h, t_s^h, \theta_s\}} \mathcal{L}_{\mathrm{align}} =  \sum_{i,j}\rho \big(\Pi(\hat{\mathbf{J}})_{j}^{i} - \mathbf{J}_{j}^{i} \big),
\end{equation}
where \(\hat{\mathbf{J}}\) denotes the 3D joint positions of the SMPL-X model, differentiably computed from the model parameters.  
\(\Pi(\cdot)_{j}^{i}\) represents the projection of the \(j\)-th joint in the \(i\)-th camera view, and \(\rho\) is the robust Geman-McClure error function~\cite{GemanMcClure1987}.

For object generation, the object $O_s$ is first initialized from the segmented object part of Anchor NeRF, and then fully completed via SDS optimization: $\nabla_{\Phi} \mathcal{L}_{\mathrm{SDS}}^O(\renderedimage)$. To preserve the overall human-object interaction, $\nabla_{\Phi} \mathcal{L}_{\mathrm{SDS}}^I(\renderedimage)$ is jointly applied during optimization. Finally, the object mesh is extracted using the marching cubes algorithm~\cite{lorensen1998marching}.

\subsection{Motion Synthesis}
\label{subsec:motion}
Following the interaction composition, this part generates the 4D HOI, \emph{i.e.} $ \{T_{\textrm{inter, (motion)}}, (H_s, O_s)\} \rightarrow \{(H_i, O_i)\}_{i=0}^{L-1}$, with anchor keypoints based prior distillation from pre-trained VDMs. We first introduce \textit{(i)} \emph{Anchor keypoints extraction}, and then \textit{(ii)} \emph{Motion optimization with anchor keypoints}.

\paragraph{Anchor keypoints extraction.}
We extract body and contact keypoints as anchor keypoints from videos generated by the video diffusion model.

\emph{ \textit{(a)} Video generation.} We adopt a VDM~\cite{zhang2025packing} to generate a video \(\{F_l^{\text{rgb}}\}_{i=0}^{L-1}\) with \(L\) frames, given the textual prompt \(T_{\text{inter}}(T_{\text{motion}})\) and the rendered static HOI image \(I^{\text{rgb}}\). 
% Notably, the first frame \(F_0^{\text{rgb}}\) matches \(I^{\text{rgb}}\).

\emph{ \textit{(b)} Body keypoints.} Since inter-subject occlusions often degrade 3D pose estimation, we instead use 2D body keypoints detected by OpenPose~\cite{cao2019openpose}, which offer robust cues even under occlusion. For each frame, OpenPose predicts 18 keypoints \((j_i, \omega_i)\), where \(j_i\) denotes the normalized pixel coordinates and \(\omega_i\) the corresponding confidence scores. The keypoints are then reordered to match the SMPL-X joint definitions.

\emph{ \textit{(c)} Contact keypoints.} While contacts typically occur in occluded regions between subjects, we extract reliable contacts based on the 3D geometric proxy underlying the generated compositional interaction. Specifically, we apply farthest point sampling to obtain a representative subset of surface points from object mesh, denoted as $\mathbf{V}_o \in \mathbb{R}^{N_o \times 3}$. The human vertices used for contact parsing are denoted as $\mathbf{C}(\mathbf{V}_h) \in \mathbb{R}^{N_h \times 3}$, where $\mathbf{C}(\cdot)$ refers to a heuristic selection of SMPL-X mesh vertices as potential contact candidates~\cite{hassan2019resolving}. To ensure the precision of contact identification, we combine geometric proximity~\cite{bhatnagar22behave, zhang2020place} and normal alignment~\cite{grady2021contactopt, yang2021cpf} constraints, and extract the valid points $\mathcal{C}_{\text{valid}}$ as contact keypoints:

% \begin{equation}
% \mathcal{C}_{\text{valid}} = \left\{ (i, j) \;\middle|\; (1 - \mathbf{n}_o^i \cdot \mathbf{n}_h^j) < \tau_n,\; \rho(\| v_o^i - v_h^j \|_2) < \tau_d \right\},
% \end{equation}

\begin{equation}
\begin{split}
\mathcal{C}_{\text{valid}} = \Big\{ (i, j) \;\Big|\; & (1 - \mathbf{n}_o^i \cdot \mathbf{n}_h^j) < \tau_n, \\
& \rho(\| v_o^i - v_h^j \|_2) < \tau_d \Big\},
\end{split}
\end{equation}

Here, $(i, j)$ denotes the constructed nearest-neighbor pairs $(i, j)$, where the $j$-th human vertex $v_j \in \mathbf{C}(\mathbf{V}_h)$ is the closest to the $i$-th object surface point $v_i \in \mathbf{V}_o$. The normal alignment term $(1 - \mathbf{n}_o^i \cdot \mathbf{n}_h^j)$ encourages the object and the human normals to be antiparallel, thereby favoring physically plausible contacts. This is consistent with the physical fact that interactions occur in contact regions where opposing forces are exerted. The geometric proximity term $\rho(\| v_o^i - v_h^j \|_2) < \tau_d$ encourages physically plausible contacts with close distance. $\rho(\cdot)$ is the Geman-McClure error function~\cite{GemanMcClure1987}.

\paragraph{Motion optimization with anchor keypoint.}

We then transfer the motion from the generated video to 3D HOI by optimizing the human and object representations
\[
\left\{\left(\{s, r_i^h, t_i^h, M(\theta_i, \Theta)\}, \{r_i^o, t_i^o, \Phi\}\right)\right\}_{i=0}^{L-1}.
\]

Among these, the motion-dependent parameters \(\{(r_i^h, t_i^h, \theta_i), (r_i^o, t_i^o)\}\) are optimized, while the motion-invariant variables \(\{s, \Theta, \Phi\}\), inherited from the static interaction, are kept fixed and shared across frames. 
% Recall that \(\theta_i\), \(r_i^h\), and \(t_i^h\) denote the human articulation pose, global rotation, and translation, respectively, while \(r_i^o\) and \(t_i^o\) represent the object's rotation and translation.

Taking anchor keypoints as tracking cues, the parameters are then optimized by minimizing the following objective:

\begin{equation}
\label{equation:total}
\mathcal{L}_{\text{total}} = \lambda_{\mathbf{J}} \mathcal{L}_{\mathbf{J}} + \lambda_{\mathbf{C}} \mathcal{L}_{\mathbf{C}} + \lambda_{\text{pen}} \mathcal{L}_{\text{pen}} + \lambda_{\text{reg}} \mathcal{L}_{\text{reg}}.
\end{equation}

\begin{equation}
\label{equation:body_key}
\mathcal{L}_{\mathbf{J}} = \frac{1}{N} \sum_{i=1}^{N} w_i  \rho \left(\hat{j}_i - j_i \right)
\end{equation}

\begin{equation}
\label{equation:contact_key}
\mathcal{L}_{\mathbf{C}} =
\frac{1}{|\mathbf{C}|}
\sum_{i \in \mathbf{C}}
\left \Vert
    p_h^i - p_o^i
\right \Vert_2^2
\end{equation}
$\mathcal{L}_{\mathbf{J}}$ minimizes the distance between the re-projected SMPL joints and the predicted body keypoints, where $w_i$ is the confidence score of each keypoint, $\rho$ is the Geman-McClure error function~\cite{GemanMcClure1987}, and $N$ is the number of keypoints. $\mathcal{L}_{\mathbf{C}}$ minimizes the distance between the $i_{th}$ paired contact keypoint of the human and the object. $\mathcal{L}_{\text{pen}}$ penalizes physical interpenetration between human and object, following ~\cite{mihajlovic2022coap}. The regularization term $\mathcal{L}_{\text{reg}}$ includes the mean squared error between the rendered model frame and video frame, a self-penetration penalty, and a temporal smoothness term. %Details are provided in the Supplementary Material.

\section{5\quad Experiments}
To evaluate the effectiveness of AnchorHOI, we conduct a comprehensive comparison with representative methods across 4D visual quality, motion fidelity, and overall interaction plausibility. In addition, we benchmark AnchorHOI against leading static 3D HOI methods to further assess its capability in 3D interaction generation. Experimental results demonstrate that AnchorHOI achieves superior performance in both dynamic 4D and static 3D HOI generation.

%%%%%%%%%%%%%%%%%%%%%%%%%%%%%%%%%%%%%%%%%%%%%%%%%%%%%%%%%%%%%%%%%
\begin{figure*}[!t]
    \centering
    \includegraphics[width=0.75\linewidth]{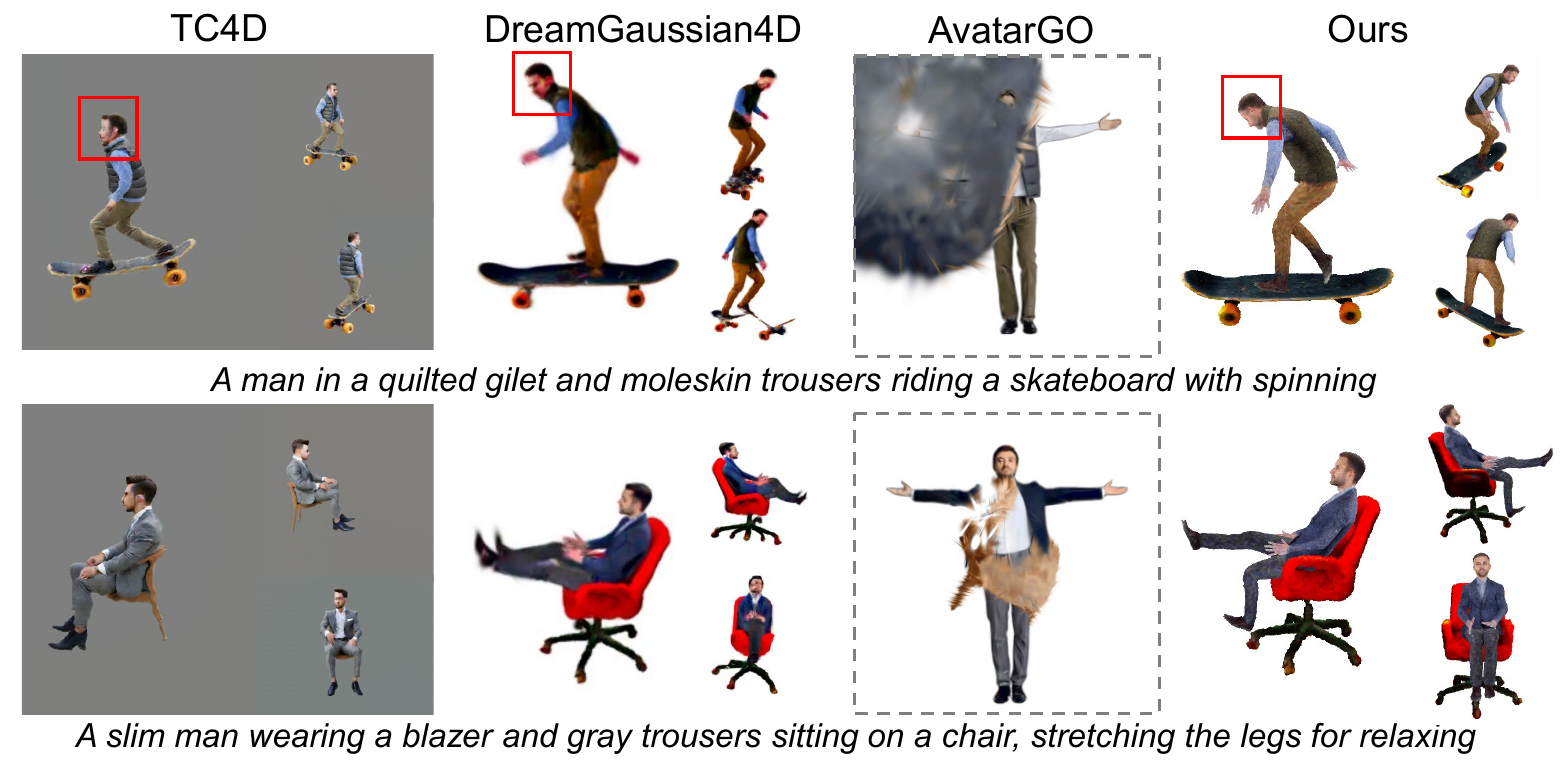}
    %\vspace{-0.5em}
    \caption{Overall 4D comparison results.}
    \label{fig:4d_comp}
    %\vspace{-0.5em}
\end{figure*}

\begin{figure*}[!t]
    \centering
    \includegraphics[width=0.75\linewidth]{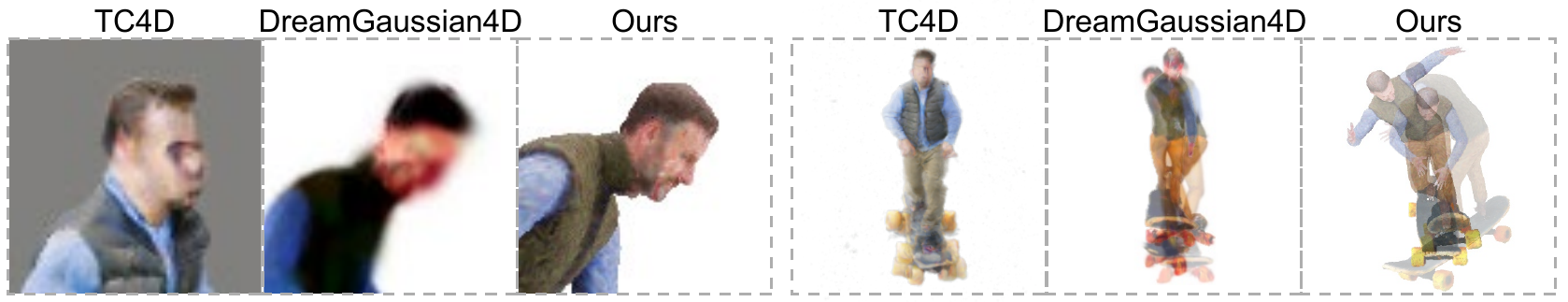}
    %\vspace{-0.5em}
    \caption{Zoom-in details (left) and motion-centric comparison (right), where opacity conveys motion amplitude: lower opacity indicates larger motion, and vice versa.}
    \label{fig:4d_detail}
    %\vspace{-0.5em}
\end{figure*}

\begin{table*}[!t]
\centering
\small
  \renewcommand{\arraystretch}{1.}
  \renewcommand{\tabcolsep}{3.pt}
  
  %\vspace{-0.5em}

\begin{tabular}{lc|cc|ccccc}
\toprule
 &   \multicolumn{1}{c}{\textbf{\textit{CLIP Score}}}   &   \multicolumn{2}{c}{\textbf{\textit{GPT-4V Selection (\%)}}}   &   \multicolumn{5}{c}{\textbf{\textit{User Studies}}}   \\
 \textbf{Methods} & Semantic $\uparrow$ & Contact $\uparrow$ & Overall $\uparrow$ & Semantic $\uparrow$ & Contact $\uparrow$ & Penetration $\uparrow$ & Motion $\uparrow$ & Overall $\uparrow$ \\
\midrule
\textbf{DreamGaussian4D} & 0.2833 & 12.50 & 25.00 & 2.339 & 2.382 & 1.594 & 2.634  & 3.339 \\
\textbf{TC4D}            & 0.3017 & 16.67 & 20.83 & 3.119 & 2.321 & 2.863 & 2.398 & 3.664 \\
\textbf{Ours}            & \textbf{0.3149} & \textbf{70.83} & \textbf{54.17} & \textbf{4.794} & \textbf{4.756} & \textbf{4.673} & \textbf{4.874} & \textbf{4.833} \\

\bottomrule
\end{tabular}
\caption{{Quantitative 4D Comparison Results.}}
\label{table:4D_results}
%\vspace{-1em}
\end{table*}

\subsection{Implementation Details}
We adopt DeepFloyd~\cite{deepfloyd-if} and a multi-view-consistent image diffusion model~\cite{shi2023MVDream} to compute SDS gradients, and use the latest video diffusion models~\cite{zhang2025packing} to generate 5-second sequences. SMPL-X and VPoser~\cite{SMPL-X:2019} serve as human body priors. OpenPose~\cite{cao2019openpose} is used to extract body joints, and Grounded-SAM~\cite{ren2024grounded} provides instance segmentation masks. We optimize using Adam~\cite{kingma2015adam} with a learning rate of 0.01, running 3{,}000 iterations for interaction composition and 1{,}000 for motion synthesis, on a NVIDIA A6000 GPU.

%on a single NVIDIA A6000 GPU. %See Supplementary Material for additional details.

\subsection{Experimental Setup}

\paragraph{Evaluation baselines.} We conduct both quantitative and qualitative evaluations to compare \nickname{} with representative 4D generation baselines. Specifically, we compare against AvatarGO~\cite{cao2024avatargo}, DreamGaussian4D~\cite{ren2023dreamgaussian4d}, and TC4D~\cite{bahmani2024tc4d}. Among them, AvatarGO is the most closely related to our approach. However, since the 4D component of AvatarGO has not been publicly released, we include only its static 3D results without motion.

\paragraph{Evaluation metrics.} To enable a thorough and reliable quantitative evaluation, we conduct perceptual studies and compute metrics from three complementary perspectives: CLIP score, GPT-4V selection, and perceptual user studies, covering principal quality dimensions such as semantic consistency, physical plausibility, and motion quality.

\subsection{Qualitative Evaluations}

\paragraph{4D HOI comparison.}
Figure~\ref{fig:4d_comp}, together with Figure~\ref{fig:4d_detail}, presents a comprehensive comparison of 4D HOI generation results.
We observe the following: (1) TC4D employs video diffusion models with SDS but treats the entire scene holistically, lacking localized or component-wise motion—an essential element for modeling human-object interactions. (2) DreamGaussian4D, even under powerful guidance (using the same VDM-generated videos as ours), struggles to animate composited HOIs due to its sole reliance on RGB and mask cues. For generating complex HOI dynamics, pixel-level cues alone are insufficient to capture meaningful interactions. (3) AvatarGO cannot be directly compared in 4D due to the unavailable code. However, its static 3D results are fundamentally limited, as it models only position without considering human articulation variance (e.g., humans remain in a fixed canonical pose) while pose variation is necessary for everyday interactions. In the chair example, although AvatarGO can optimize the object position for interaction, the absence of pose modeling prevents it from producing plausible results. Since its 4D outcome directly depends on the quality of static 3D composition, limitations in 3D modeling inevitably result in implausible 4D outputs.

In contrast, our method achieves more realistic details (e.g., finer visual results in Figure~\ref{fig:4d_detail}(left)), more expressive motion (e.g., larger amplitudes in the skateboard example of Figure~\ref{fig:4d_detail}(right)), and consistent contact awareness throughout the 4D sequences.

\begin{figure}[!t]
    %\cent
    %\vspace{0.7em}
    \centering
    \includegraphics[width=0.81\linewidth]{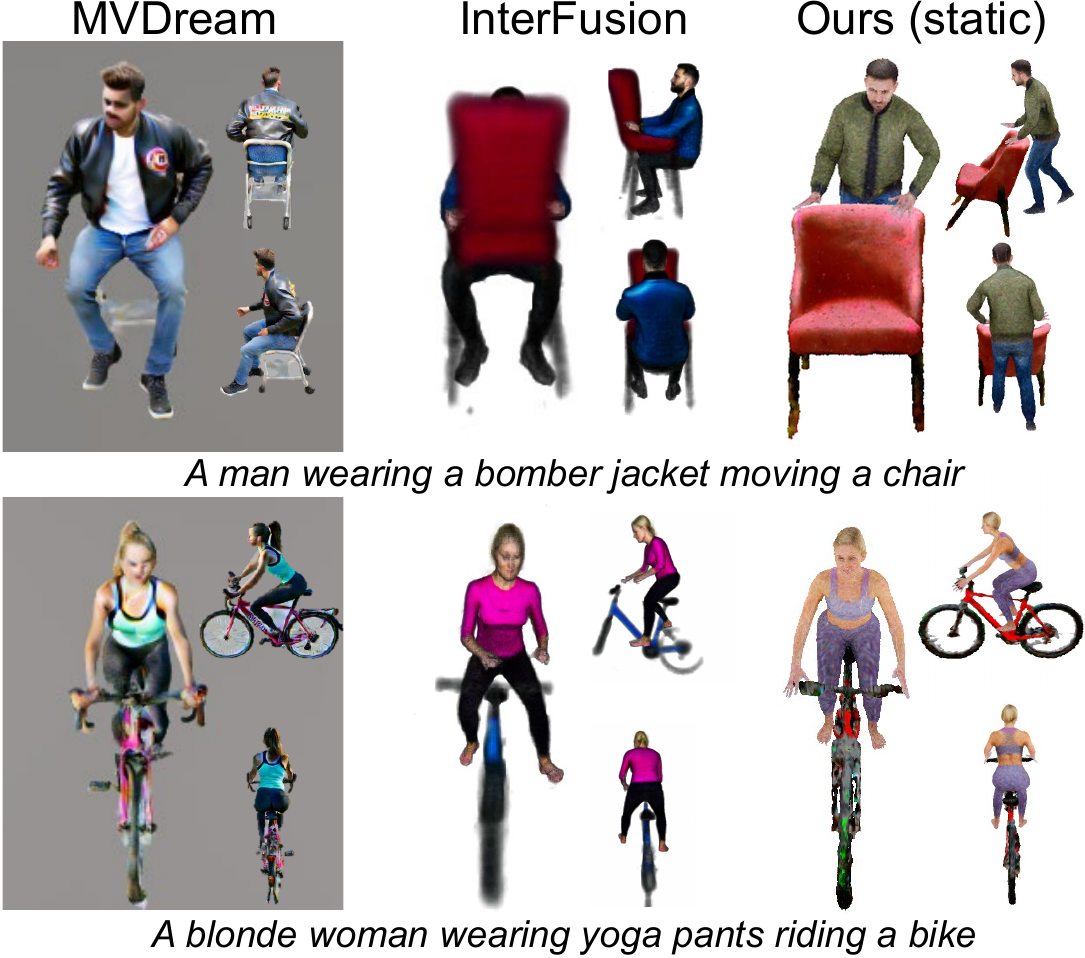}
    \caption{Qualitative 3D comparison results.}
    %\vspace{-1em}
    \label{fig:3d_comparison}
    %\vspace{-1em}
\end{figure}

\begin{figure*}[t]
  \centering
  \begin{minipage}[t]{0.49\textwidth}
    \centering
    \includegraphics[width=0.85\linewidth]{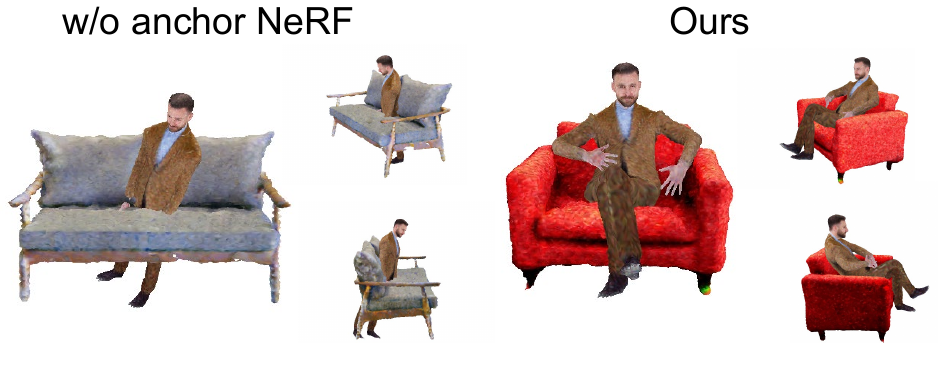}
    \caption*{\textbf{(a)} \textit{A man with sideburns in a corduroy blazer sitting on a sofa.}}
    \label{fig:3d_ablation}
  \end{minipage}
  \hfill
  \begin{minipage}[t]{0.49\textwidth}
    \centering
    \includegraphics[width=0.9\linewidth]{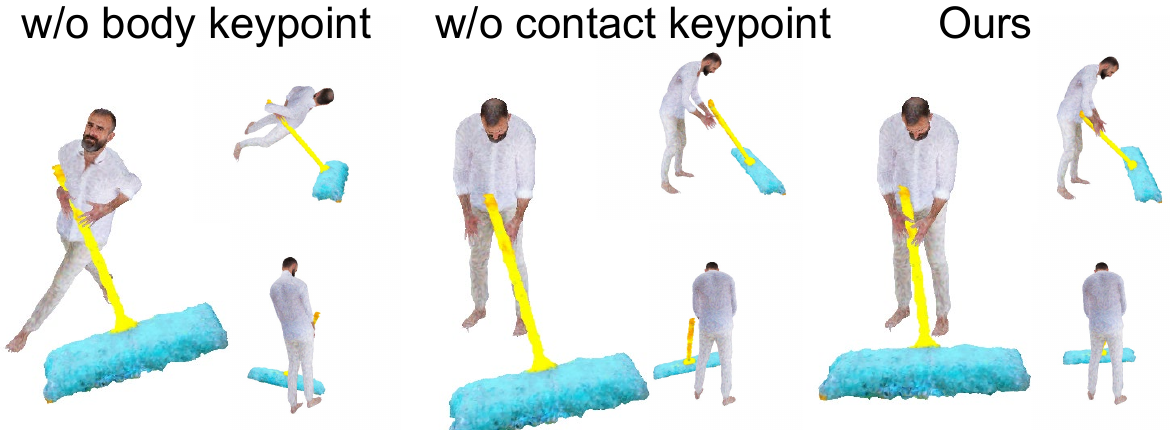}
    \caption*{\textbf{(b)} \textit{A Mediterranean man cleaning with a mop.}}
    \label{fig:4d_ablation}
  \end{minipage}
  %\vspace{-0.5em}
  \caption{Ablation studies on anchor NeRF and anchor keypoints.}
  %\vspace{-1.5em}
  \label{fig:ablation}
\end{figure*}

\paragraph{3D HOI comparison.} Although AnchorHOI targets 4D interaction generation, it also achieves superior performance in static 3D HOI generation. We compare our method with two state-of-the-art baselines, MVDream and InterFusion, as shown in Figure \ref{fig:3d_comparison}. Specifically: (1) MVDream entangles human and object representations, often causing structural artifacts, blurry contacts, or semantically invalid interactions; (2) InterFusion uses fixed, retrieved poses from image reconstructions, lacking adaptability to specific interaction contexts. Benefiting from the proposed anchor NeRF, AnchorHOI produces the most realistic and pose-adaptable 3D human-object interactions, effectively capturing complex contacts across diverse scenarios. Numeric results in Table~\ref{table:3D_results}, including CLIP scores and GPT-4V selections, further demonstrate our superiority.

%%%%%%%%%%%%%%%%%%%%%%%%%%%%%%%%%%%%%%%%%%%%%%%%%%%%%%%%%%%%%%%%%
\subsection{Quantitative Evaluations}

\paragraph{CLIP score.} Following common practice, we compute the CLIP score~\cite{radford2021clip} to measure the similarity between input text prompts and the corresponding generated results, where a higher score indicates better alignment with input descriptions. \nickname{} achieves the highest mean CLIP score across evaluation prompts in both 3D and 4D.

\paragraph{GPT-4V selection.} Following InterFusion, we further leverage the advanced image understanding capabilities of GPT-4V to enable a more fine-grained evaluation. Specifically, we prompt GPT-4V to (1) select the overall preferred result based on interaction criteria, and (2) assess contact accuracy as a physically grounded metric in isolation. No in-context examples are provided during prompting.

\paragraph{User studies.} We further conduct perceptual user studies for 4D evaluation, following \cite{bylinskii2023towards}. The numerical results are reported in Table~\ref{table:4D_results}.

%Details are provided in the Supplementary Material, and the numerical results are reported in Table~\ref{table:4D_results}.

\begin{table}[!t]
\centering
\small
\renewcommand{\arraystretch}{1.}
\setlength{\tabcolsep}{3pt}

%\vspace{-0.5em}

\begin{tabular}{lc|cc}
\toprule
 & \multicolumn{1}{c}{\textbf{\textit{CLIP Score}}}
 & \multicolumn{2}{c}{\textbf{\textit{GPT-4V Selection (\%)}}} \\
\textbf{Methods} & Semantic $\uparrow$ & Contact $\uparrow$ & Overall $\uparrow$ \\
\midrule
\textbf{MVDream}           & 0.2948 & 15.79 & 10.53 \\
\textbf{InterFusion}       & 0.2951 & 21.05 & 26.32 \\
\textbf{AvatarGO (static)} & 0.2615 & 5.26  & 10.53 \\
\textbf{Ours (static)}     & \textbf{0.3173} & \textbf{57.89} & \textbf{52.63} \\
\bottomrule
\end{tabular}
\caption{Quantitative 3D comparison results.}
\label{table:3D_results}
%\vspace{-1.5em}
\end{table}

%%%%%%%%%%%%%%%%%%%%%%%%%%%%%%%%%%%%%%%%%%%%%%%%%%%%%%%%%%%%%%%%%
\subsection{Ablation Study}
\paragraph{Ablation on anchor NeRF.} We conduct ablations with and without anchor-NeRF. In the w/o Anchor-NeRF setting, the object and human are generated separately and then composed via SDS directly. As shown in Figure~\ref{fig:ablation} (a), the human pose fails to converge, as position rather than articulation is optimized to satisfy the “sitting on” interaction, resulting in unrealistic results. In contrast, Anchor-NeRF effectively bridges SDS gradients to human articulation, enabling stable convergence and even complex poses (e.g., a crossed-legs posture). This example, along with the quantitative results in Table~\ref{table:ablation_gpt4v} (up), demonstrates the effectiveness of Anchor-NeRF for faithful interaction composition.

\paragraph{Ablation on anchor keypoints.} Figure~\ref{fig:ablation} (b) illustrates the role of anchor keypoints. Body keypoints are essential for capturing human pose; without them, the resulting postures are implausible, even maintaining interaction contacts. Without contact keypoints, the generated results may be visually plausible but lack meaningful physical contact. In contrast, our full setting with both body and contact keypoints produces realistic 4D interaction sequences, as further supported by the quantitative results in Table~\ref{table:ablation_gpt4v} (down).

\begin{table}[!t]
\centering
\small
\renewcommand{\arraystretch}{1.1}
\setlength{\tabcolsep}{5pt}

%\vspace{-0.5em}

\begin{tabular}{lc}
\toprule
\textbf{Variants} & {\textbf{\textit{GPT-4V Selection (\%)}}}  \\
\midrule
\textbf{w/o anchor NeRF}    & 5.88  \\
\textbf{Ours (static)}      & \textbf{94.12}  \\
\toprule
\textbf{w/o body keypoints}    & 5.89  \\
\textbf{w/o contact keypoints} & 17.65 \\
\textbf{Ours} & \textbf{76.47}\\
\bottomrule
\end{tabular}
\caption{Quantitative ablation results by GPT-4V.}
\label{table:ablation_gpt4v}
%\vspace{-1.5em}
\end{table}

\section{6\quad Conclusion}

In this paper, we presented AnchorHOI, a novel framework for zero-shot 4D human-object interaction (HOI) generation with an anchor-based prior distillation strategy. Our approach takes a step forward in interaction-aware 4D generation by effectively leveraging hybrid priors from both image and video diffusion models. Specifically, AnchorHOI tailors anchor NeRFs for interaction composition and anchor keypoints for motion synthesis, enabling effective and reliable prior distillation. Experimental results show that AnchorHOI achieves state-of-the-art performance in both static 3D and dynamic 4D HOI generation.

\paragraph{Limitations and Future Work.}

% Our method still exhibits several limitations and open up interesting avenues for future research. One limitation is the current assumption of continuous contact between humans and objects, which overlooks the dynamic nature of contact in real-world scenarios. Fine-grained temporal contact modeling could enable more generalizable representations of interaction. Another limitation is that our method is currently designed for rigid objects. Extending the framework to accommodate articulated objects with kinematic properties, such as washers, microwaves, and other commonly manipulated objects, would be a meaningful step toward practical applications.

One limitation of our method is the current assumption of continuous contact between humans and objects, which overlooks the dynamic nature of contact in real-world scenarios. Another limitation is that our method is currently designed for rigid objects. Extending the framework to accommodate articulated objects with kinematic properties would be a meaningful step toward practical applications.

\section{Acknowledgments}
This work was supported in part by the NSFC (62325211, 62132021) and the Major Program of Xiangjiang Laboratory (23XJ01009).

\bibliography{aaai2026}

\end{document}